\documentclass[journal]{IEEEtran}

\usepackage{amsfonts,dsfont}
\usepackage{balance}
\usepackage{booktabs}
\usepackage{amsmath}
\usepackage{tabularx}
\usepackage{subfigure}
\usepackage{graphicx}
\usepackage{multirow}
\usepackage{rotating}
\usepackage{algorithm}
\usepackage{algorithmic}
\usepackage{url}
\usepackage{amssymb}
\usepackage{amsthm}
\usepackage{enumerate}

\usepackage{color}

\newtheorem{mythm}{Theorem}

\begin{document}
%
\title{Adaptively-weighted Integral Space for Fast Multiview Clustering}

\author{
		Man-Sheng Chen,
		Tuo Liu,
        Chang-Dong~Wang,~\IEEEmembership{Member,~IEEE,}
        Dong~Huang,~\IEEEmembership{Member,~IEEE,}\\
        Jian-Huang Lai,~\IEEEmembership{Senior Member,~IEEE}
}


\maketitle

\begin{abstract}
Multiview clustering has been extensively studied to take advantage of multi-source information to improve the clustering performance. In general, most of the existing works typically compute an $n\times n$ affinity graph by some similarity/distance metrics (e.g. the Euclidean distance) or learned representations, and explore the pairwise correlations across views. But unfortunately, a quadratic or even cubic complexity is often needed, bringing about difficulty in clustering large-scale datasets. Some efforts have been made recently to capture data distribution in multiple views by selecting view-wise anchor representations with $k$-means, or by direct matrix factorization on the original observations. Despite the significant success, few of them have considered the view-insufficiency issue, implicitly holding the assumption that each individual view is sufficient to recover the cluster structure. Moreover, the latent integral space as well as the shared cluster structure from multiple insufficient views is not able to be simultaneously discovered. In view of this, we propose an \underline{A}daptively-weighted \underline{I}ntegral Space for Fast \underline{M}ultiview \underline{C}lustering (AIMC) with nearly linear complexity. Specifically, view generation models are designed to reconstruct the view observations from the latent integral space with diverse adaptive contributions. Meanwhile, a centroid representation with orthogonality constraint and cluster partition are seamlessly constructed to approximate the latent integral space. An alternate minimizing algorithm is developed to solve the optimization problem, which is proved to have linear time complexity \textit{w.r.t.} the sample size. Extensive experiments conducted on several real-world datasets confirm the superiority of the proposed AIMC method compared with the state-of-the-art methods.
\end{abstract}


%
\IEEEpeerreviewmaketitle

\section{Introduction}

In the big data era, a large quantity of multimedia data from Internet as well as social media spring up, where they are often collected without the label information~\cite{DBLP:journals/tkde/WangYL20,DBLP:conf/mm/Huang0P021,DBLP:conf/mm/ZhangLW0DZ21}. Clustering techniques have attracted more and more attention in the unsupervised data mining and machine learning community. Many classical clustering methods have been proposed to automatically explore the correlations between different data objects, such as $k$-means as well as its variants, spectral clustering, and subspace clustering. Despite impressive performance, they are merely suitable to single view scenario. In practical applications, data are often represented with diverse features from multiple sources or feature extractors. For instance, a webpage is represented by multiview features based on text, image and video. News can be reported by different languages. Multiple views or feature subsets comprehensively describe different aspects of the data, and it is generally infeasible to just consider single view rather than multiple views. Therefore, multiview clustering is developed to well integrate multiple views or subspace structures to improve the clustering performance.

Considerable efforts have been recently made in the development of multiview clustering to investigate the diverse and complementary information among multiple views~\cite{DBLP:conf/ijcai/NieLL17,DBLP:conf/mm/ZhangLW0DZ21,DBLP:conf/mm/ZhangWLZZLZZ21}, in which the existing literatures can be divided into three main categories, i.e., the co-training based models, graph based algorithms and subspace clustering based methods. The co-training based models aim at maximizing the mutual agreement between different views via alternately training~\cite{DBLP:conf/nips/KumarRD11,DBLP:conf/icml/KumarD11}. For instance, in~\cite{DBLP:conf/nips/KumarRD11}, Kumar et al. co-regularized the clustering hypotheses such that different representations could agree with each other. The work in~\cite{DBLP:conf/icml/KumarD11} proposed to seek for the clusters agreeing across multiple views. Ye at al. studied the underlying clustering by maximizing the sum of weighted affinities between different clusterings of multiple views~\cite{DBLP:conf/icpr/YeLYZ16}. The graph based algorithms attempt to combine multiple views for recovering the relationships between views~\cite{DBLP:conf/aaai/NieCL17,ZhangOne-step20}. One of the earliest literatures made bipartite graph construction to link features from different views~\cite{de2005spectral}. In~\cite{DBLP:conf/aaai/NieCL17}, Nie et al. proposed to simultaneously learn the clustering/semi-supervised classification as well as local structure. The work in~\cite{DBLP:journals/tkde/WangYL20} coupled the similarity-induced graph associated with multiple views, unified graph and indicator into a unified framework. Supposing that views are generated from the latent subspace, the subspace clustering based methods are developed to learn subspace representations based on the self-expressive property. In~\cite{luo2018consistent}, Luo et al. simultaneously explored consistency and specificity in multiple subspace representations. The work in~\cite{ZhangOne-step20} proposed a one-step kernel subspace clustering model, in which the shared affinity representation was learned from multiple views. Chen et al. jointly studied the low-rank representation tensor as well as affinity matrix while keeping the local structures~\cite{DBLP:journals/pr/ChenXZ20}. Nevertheless, most of existing works suffer from high computational cost (typically quadratic or even cubic complexity for inevitable $n\times n$ affinity graph construction), which seriously restricts their efficiency when dealing with large-scale multiview data.

\begin{figure*}[!t]
	\begin{center}
		\includegraphics[width=1.0\linewidth]{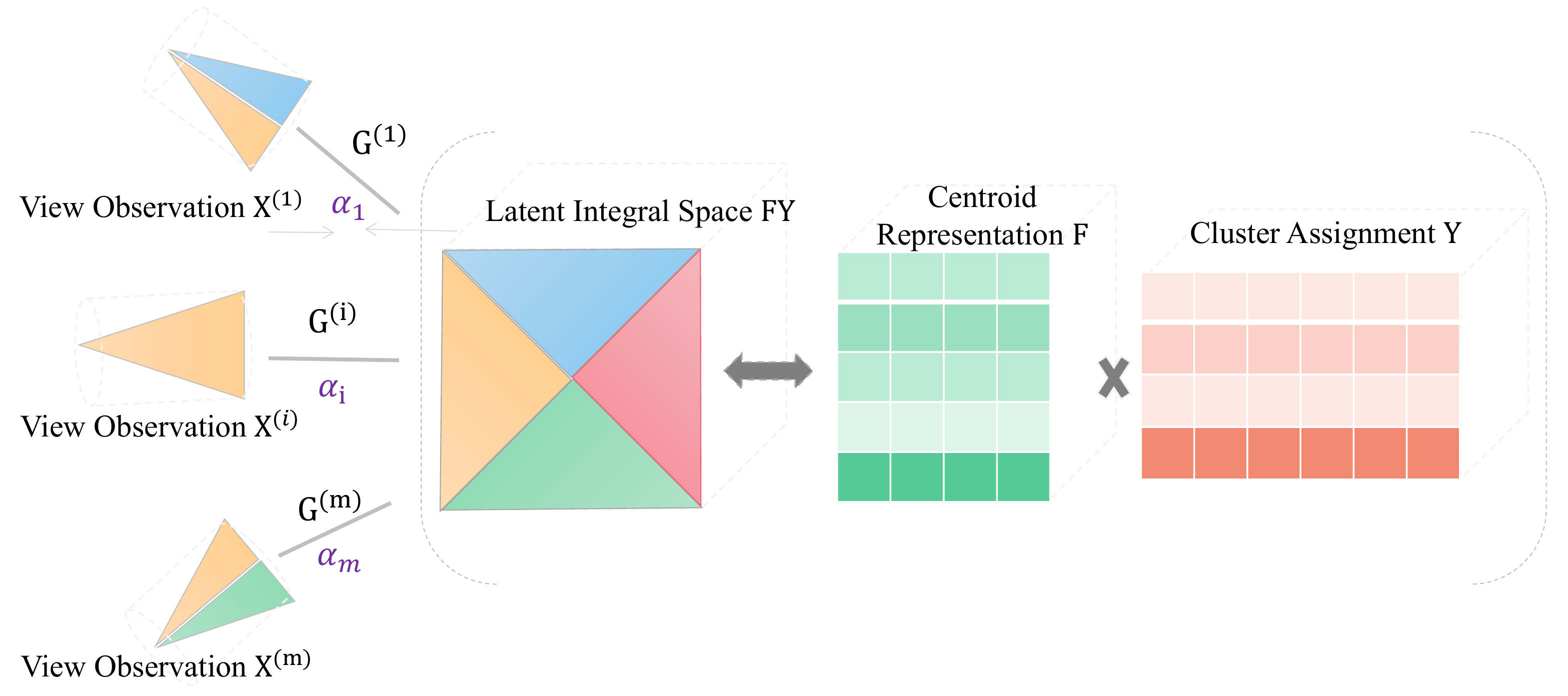}
		\caption{Illustration of the proposed AIMC method. The adaptively-weighted latent integral space depicting the common cluster assignment is modeled via view generation models with input view observations carrying diverse adaptive confidences.}
		\label{fig:Flowchart}
	\end{center}
\end{figure*}

In view of this, some large-scale oriented multiview clustering algorithms have been proposed, which can be mainly divided into two categories, i.e., matrix factorization style algorithms~\cite{DBLP:conf/ijcai/CaiNH13a,DBLP:journals/pr/NieSL20,DBLP:journals/tip/YangZNWYW21,DBLP:conf/iccv/0003LYL0LS21} and anchor graph based methods~\cite{DBLP:conf/aaai/LiNHH15,DBLP:conf/aaai/KangZZSHX20}. The matrix factorization style algorithms are able to reduce computational burden without the construction of affinity graph. For instance, Wang et al. attempted to explore the cluster structure with a constrained factor matrix~\cite{DBLP:conf/ijcai/WangNHM11}. The work in~\cite{DBLP:conf/aaai/HanSNL17} treated the intermediate factor matrix as a diagonal matrix in 3-factor factorization, improving the efficiency of matrix multiplication in optimization. In~\cite{DBLP:conf/ijcai/CaiNH13a}, Cai et al. efficiently combined heterogeneous factor matrices of large-scale data. Nie et al. ~\cite{DBLP:journals/pr/NieSL20} proposed to conduct the clustering task simultaneously on row and column of the original data. In~\cite{DBLP:conf/iccv/0003LYL0LS21}, Liu et al. unified matrix factorization with partition learning without the non-negativity constraint. However, the efficiency of this kind of algorithms would significantly decrease when facing with large data feature dimension, owing to their direct factorization of the original data observation. The anchor graph based methods often deploy correlations between anchor representation and original data to construct the anchor graph. For instances, the work in~\cite{DBLP:conf/aaai/LiNHH15} achieved the fusion graph construction for multiple anchor graphs via a local manifold fusion strategy. Kang at al. developed an anchor graph-based subspace clustering framework~\cite{DBLP:conf/aaai/KangZZSHX20}, in which view-specific graph was explored with pre-defined anchor representations. The work in~\cite{DBLP:conf/mm/SunZWZTLZW21} attempted to explore anchor graph construction based on the underlying data distribution. Similarly, Wang et al. studied the shared anchor graph with the guidance of consensus anchor representation in multiview data~\cite{DBLP:journals/tip/WangLZZZGZ22}. Despite significant success, the view-insufficiency issue is neglected among them, with the assumption that each individual view is sufficient to recover the cluster partition. Besides, they fail to simultaneously discover the latent integral space from multiple insufficient views as well as the common cluster structure shared by multiview data.

To address the aforementioned problems, in this paper, a novel \underline{A}daptively-weighted \underline{I}ntegral Space for Fast \underline{M}ultiview \underline{C}lustering (AIMC) method is developed. Instead of performing direct factorization of the original data observation as the existing matrix factorization based literatures, a latent integral space revealing the cluster structure is focused by considering different adaptive contributions from multiple views in our framework, which is the enhanced formulation associated with the general fast multiview $k$-means and flexibly avoids the curse of dimensionality. To be specific, for each view, we design a view generation model to map the latent integral space into the corresponding view observation while adpatively considering their diverse contributions. Reversely, with input view observations, we can restore the latent integral space by mapping back from given view observations with the corresponding self-conducted confidence. Additionally, a centroid representation with orthogonality constraint and cluster assignment are directly constructed to approximate the latent integral space, which not only focuses on the space itself but also seamlessly associates it with the underlying data partition. For clarity, the flowchart of the proposed AIMC method is illustrated in \figurename~\ref{fig:Flowchart}. A four-step alternate minimizing strategy is designed with empirical convergence, in which each variable can be updated efficiently to receive the closed-form solution. Consequently, the reconstruction error and distortion error of data partition are merged in our model to further improve the effectiveness and efficiency of clustering.

The main contributions of our work are summarized as follows:
\begin{itemize}
	\item Considering the view-insufficiency issue, we study an adaptively weighted integral space for fast multiview clustering, which is coupled with the underlying cluster partition.
	\item We develop an alternate minimizing strategy to solve the optimization problem, by which the proposed method is proved to have linear time complexity	\textit{w.r.t.} the number of samples.
	\item Extensive experiments are conducted on multiple challenging datasets, including several big datasets, to demonstrate the superiority of the proposed method compared with the state-of-the-art methods.
\end{itemize}

The rest of this paper is organized as follows. In Section~\ref{sec:preliminaries}, the main notations and basic preliminaries of matrix factorization are provided. The proposed Adaptively-weighted Integral Space for Fast Multiview Clustering method is described in Section~\ref{sec:proposed method} in which the optimization algorithm and time computational analysis as well as space complexity analysis are provided. In Section~\ref{sec:experiments}, the experimental results are reported, where nine real-world datasets are used and nine state-of-the-art methods are compared. Finally, the paper is concluded in Section~\ref{sec:conclusion}.

\section{Preliminaries}
\label{sec:preliminaries}
\subsection{Notations}

Throughout this paper, matrices are represented as uppercase letters. The $i$-th row of matrix $A$ is denoted as $A_{i,:}$, with its $j$-th entry being $A_{ij}$. $\|A\|_F^2$ represents the square of Frobenius norm of matrix $A$, and $\|A_{i,:}\|_1$ denotes the $l_1$ norm of the row vector $A_{i,:}$. The trace of $A$ can be written as $Tr(A)$. We denote the $v$-th view observation of $A$ as $A^{(v)}$. In particular, $I_d$ stands for the $d$ dimensional identity matrix.

\subsection{Matrix Factorization}

Non-negative matrix factorization (NMF) and its variants are often employed in clustering owing to the success in feature extraction. In this part, we will make brief introduction of NMF and one-sided orthogonal NMF~\cite{Ding2006OrthogonalNM}. Given the data matrix $X\in\mathbb{R}^{d\times n}$ drawn from $k$ partitions, the regularization term about matrix factorization can be generally defined as
\begin{equation}
	\begin{aligned}
		\label{mf1}
		\mathop{\min}\mathcal{L}\left(X,WH\right) ,~~~\text{s.t.~~~}W\in\mathcal{C}_1,H\in\mathcal{C}_2,
	\end{aligned}
\end{equation}
where $\mathcal{L}\left(\cdot\right)$ stands for the loss function. $W$ and $H$ are the factor matrices constrained respectively by $\mathcal{C}_1$ and $\mathcal{C}_2$.

NMF aims to decompose a non-negative input matrix into two non-negative representation products, i.e., the base matrix and coefficient representation, whose decomposition can be formulated as
\begin{equation}
	\begin{aligned}
		\label{nmf1}
		\mathop{\min}\limits_{W,H}\|X_+-WH\|_F^2,~~~\text{s.t.~~~}W\geq0,H\geq0,
	\end{aligned}
\end{equation}
where $W\in\mathbb{R}^{d\times k}$ is denoted as the base matrix, and $H\in\mathbb{R}^{k\times n}$ represents the coefficient representation. It can be known that $H$ is a low-rank coefficient matrix, on which the final cluster partition can be acquired by performing classical clustering algorithm (e.g. $k$-means).

The concept of one-sided orthogonal NMF was first proposed by Ding et al, in which the orthogonality constraint was applied to a factor matrix~\cite{Ding2006OrthogonalNM}. The objective formulation can be written as
\begin{equation}
	\begin{aligned}
		\label{onmf1}
		\mathop{\min}\limits_{W,H}\|X_+-WH\|_F^2,~~~\text{s.t.~~~}WW^T=I_d,H\geq0,
	\end{aligned}
\end{equation}
where the base matrix $W$ becomes an orthogonal matrix. Meanwhile, the formulation above is inherently associated with $k$-means algorithm while keeping the orthogonality constraint.

\section{The Proposed Method}
\label{sec:proposed method}

In this section, we will first describe the proposed Adaptively-weighted Integral Space for Fast Multiview Clustering (AIMC) model, followed by the detailed optimization procedure. In the meantime, an analysis about the computational time and memory usage is conducted to demonstrate the time and space efficiency of AIMC.

\subsection{Formulation}

Since each distinct view representation only consists of partial information about multiple data samples, and the data collection may be noisy, in this paper, we develop a fast multiview clustering framework based on the adaptively-weighted integral space, which depicts the data structure in essence and reveals the common cluster partition shared by different views. As shown in~\cite{DBLP:journals/pami/XuTX15}, the view specific observation can be generated by mapping from one latent integral space via certain view generation model. Given a multiview dataset $\left\{X^{(v)}\in\mathbb{R}^{d_v\times n}\right\}_{v=1}^{m}$ with its $v$-th view dimensionality, sample size and view number being respectively $d_v$, $n$ and $m$, the corresponding objective function can be formulated as follow
\begin{equation}
	\begin{aligned}
		\label{obj1}
		&\mathop{\min}\limits_{G^{(v)},F,Y}\sum_{v=1}^{m}\|X^{(v)}-G^{(v)}FY\|_F^2,\\
		&\text{s.t.~~~}{G^{(v)}}^TG^{(v)}=I_d,F^TF=I_k,Y\in Ind,
	\end{aligned}
\end{equation}
where $G^{(v)}\in\mathbb{R}^{d_v\times d}$ represents the $v$-th view generation model. $F\in\mathbb{R}^{d\times k}$ denotes the centroid representation with its column being the cluster centroid of different partitions, and $Y\in\mathbb{R}^{k\times n}$ is the cluster assignment matrix with $Y_{ij}=1$ if the $j$-th instance is assigned to the $i$-th cluster and 0 otherwise. Accordingly, the centroid representation and cluster assignment matrix are constructed to approximate the latent integral space, directly ensuring the reliability of view generation model from raw data observation to the underlying data structure. The orthogonality constraint imposed on $G^{(v)}$ aims to avoid pushing the integral space arbitrarily, while the one about $F$ attempts to attain mutual independence between different cluster centroid representation.

Despite of this, the model in Eq.~\eqref{obj1} treats each view observation equally to discover the integral space, paying no attention to different contributions of multiple views and may resulting in the sub-optimal performance. Therefore, in this paper, an adaptively weighting strategy is designed~\cite{DBLP:conf/ijcai/NieLL17}, and the overall objective function can be rewritten as,
\begin{equation}
	\begin{aligned}
		\label{obj2}
		&\mathop{\min}\limits_{G^{(v)},F,Y}\sum_{v=1}^{m}\alpha_v\|X^{(v)}-G^{(v)}FY\|_F^2,\\
		&\text{s.t.~~~}{G^{(v)}}^TG^{(v)}=I_d,F^TF=I_k,Y\in Ind,
	\end{aligned}
\end{equation}
where $\alpha_v$ is defined as
\begin{equation}
	\begin{aligned}
		\label{alphaobj}
		\alpha_v\overset{\underset{\mathrm{def}}{}}{=}\frac{1}{2\|X^{(v)}-G^{(v)}FY\|_F}.
	\end{aligned}
\end{equation}

Note that according to Eq.~\eqref{alphaobj}, $\alpha_v$ is dependent on $G^{(v)}$, $F$ and $Y$. Particularly, if view $v$ is good, $\|X^{(v)}-G^{(v)}FY\|_F$ should be small, and thus the weight $\alpha_v$ of view $v$ would be large according to Eq.~\eqref{alphaobj}. Otherwise, small weights would be assigned to poor views, which demonstrates the significance of the self-weighted learning strategy. Further, it is easy to observe that Eq.~\eqref{obj2} is equivalent to the following formulation
\begin{equation}
	\begin{aligned}
		\label{extendobj}
		&\mathop{\min}\limits_{G^{(v)},F,Y}\sum_{v=1}^{m}\|X^{(v)}-G^{(v)}FY\|_F,\\
		&\text{s.t.~~~}{G^{(v)}}^TG^{(v)}=I_d,F^TF=I_k,Y\in Ind.
	\end{aligned}
\end{equation}

To some extent, the formulation is a multiview least-absolute residual model, which can induce robustness via mitigating the impact of outliers with the present first power residual and achieving dimension reduction by flexible and lower dimension of view generation model during the learning procedure.

\subsection{Optimization}

Due to the non-smoothness and unsupervised learning, it is difficult to handle the problem in Eq.~\eqref{extendobj}. Hence, the following observation is put forward to make it tractable.
\begin{mythm}
	\label{def:objthm}
	The model in Eq.~\eqref{extendobj} is equivalent to the following formulation
	\begin{equation}
		\begin{aligned}
			\label{overallobj}
			&\mathop{\min}\limits_{G^{(v)},F,Y,\alpha_v}\sum_{v=1}^{m}\alpha_v\|X^{(v)}-G^{(v)}FY\|_F^2,\\
			&\text{s.t.~~~}{G^{(v)}}^TG^{(v)}=I_d,F^TF=I_k,Y\in Ind,\alpha\in\Delta_v.
		\end{aligned}
	\end{equation}
\end{mythm}

Then, an alternate minimizing algorithm is designed to solve the optimization problem in Eq.~\eqref{overallobj} by optimizing each variable while fixing the others.

\textbf{$G^{(v)}$-subproblem:} By fixing other variables, the optimization problem for updating $G^{(v)}$ can be formulated as
\begin{equation}
	\begin{aligned}
		\label{g1}
		\mathop{\min}\limits_{G^{(v)}}\sum_{v=1}^{m}\alpha_v\|X^{(v)}-G^{(v)}FY\|_F^2,~~~\text{s.t.~~~}{G^{(v)}}^TG^{(v)}=I_d.
	\end{aligned}
\end{equation}

Since each $G^{(v)}$ is independent from each other in terms of different views, we can further transform the optimization problem in Eq.~\eqref{g1} into the following model,
\begin{equation}
	\begin{aligned}
		\label{g2}
		\mathop{\max}\limits_{G^{(v)}}Tr({G^{(v)}}^TH^{(v)}),~~~\text{s.t.~~~}{G^{(v)}}^TG^{(v)}=I_d,
	\end{aligned}
\end{equation}
where $H^{(v)}=X^{(v)}Y^TF^T$. The optimal solution of $G^{(v)}$ can be efficiently solved by singular value decomposition (SVD) technique based on $H^{(v)}$, and specifically $G^{(v)}$ can be obtained by calculating $U_HV_H^T$, where $H^{(v)}=U_{H}\Sigma_HV_H^T$~\cite{DBLP:conf/ijcai/WangLZTLHXY19}.

\begin{algorithm}[!t]
	\caption{Adaptively-weighted Integral Space for Fast Multiview Clustering}
	\label{Alg:AIalg}
	{\bfseries Input:} Multiview dataset $\left\{X^{(v)}\in\mathbb{R}^{d_v\times n}\right\}_{v=1}^{m}$, dimension $d$ and cluster number $k$.
	\begin{algorithmic}[1]
		\STATE Initialize $G^{(v)}$, $F$, $Y$. Initialize $\alpha_v$ with $\frac{1}{m}$.
		\WHILE {not converged}
		\STATE Update $G^{(v)}, \forall v$ by solving the problem in Eq.~\eqref{g2}.
		\STATE Update $F$ by solving the problem in Eq.~\eqref{f2}.
		\STATE Update $Y$ via the optimal row obtained by Eq.~\eqref{y3}.
		\STATE Update $\alpha_v$ by solving the problem in Eq.~\eqref{alphaobj}.
		\ENDWHILE
	\end{algorithmic}
	{\bfseries Output:} The final consensus cluster assignment matrix $Y$.
\end{algorithm}

\textbf{$F$-subproblem:} By fixing the other variables, the objective function \textit{w.r.t.} $F$ can be transformed into the following formulation
\begin{equation}
	\begin{aligned}
		\label{f1}
		\mathop{\min}\limits_{F}\sum_{v=1}^{m}\alpha_v\|X^{(v)}-G^{(v)}FY\|_F^2,~~~\text{s.t.~~~}F^TF=I_k.
	\end{aligned}
\end{equation}

Similar to $G^{(v)}$, it is equivalent to solving $F$ by the following problem
\begin{equation}
	\begin{aligned}
		\label{f2}
		\mathop{\max}\limits_{F}Tr(F^TJ),~~~\text{s.t.~~~}F^TF=I_k,
	\end{aligned}
\end{equation}
where $J=\sum_{v=1}^{m}\alpha_v{G^{(v)}}^TX^{(v)}Y^T$. Therefore, the closed-form solution of $F$ is equal to $U_JV_J^T$, where $J=U_J\Sigma_JV_J^T$.

\textbf{$Y$-subproblem:} By fixing other variables, $Y$ can be updated by solving the following problem
\begin{equation}
	\begin{aligned}
		\label{y1}
		&\mathop{\min}\limits_{Y}\sum_{v=1}^{m}\alpha_v\|X^{(v)}-G^{(v)}FY\|_F^2,\\
		&\text{s.t.~~~}Y_{ij}\in\left\{0,1\right\},\sum_{i=1}^{k}Y_{ij}=1,\forall j=1,2,\cdots,n.
	\end{aligned}
\end{equation}

However, it is difficult to directly solve $Y$ as a whole, since there is one and only one non-zero entry (i.e., 1) in each column of the cluster assignment matrix. Inspired by~\cite{DBLP:conf/sdm/DingH05}, we can solve the optimization problem in Eq.~\eqref{y1} independently for each object. Therefore, for each object, we can have
\begin{equation}
	\begin{aligned}
		\label{y2}
		&\mathop{\min}\limits_{Y_{:,j}}\sum_{v=1}^{m}\alpha_v\|X_{:,j}^{(v)}-G^{(v)}FY_{:,j}\|^2,\\
		&\text{s.t.~~~}Y_{:,j}\in\left\{0,1\right\}^k,\|Y_{:,j}\|_1=1.
	\end{aligned}
\end{equation}

We can settle the above optimization problem by finding one optimal row $i^{\ast}$ in $Y_{:,j}$ such that $Y_{i^{\ast},j}=1$ and 0 for other entries in the $j$-th column. Then, the optimal row can be searched by
\begin{equation}
	\begin{aligned}
		\label{y3}
		i^{\ast}=\mathop{\arg\min}_{i}\sum_{v=1}^{m}\alpha_v\|X_{:,j}^{(v)}-G^{(v)}F_{:,i}\|^2.
	\end{aligned}
\end{equation}

As a matter of fact, the optimal solution of cluster assignment can be obtained by minimizing the distance between the data sample and the cluster centroid.

\textbf{$\alpha_v$-subproblem:} By fixing the other variables except $\alpha_v$, the optimization problem is equivalent to dealing with the model in Eq.~\eqref{alphaobj}, which is solved by the computed $G^{(v)}$, $F$ and $Y$ in each iteration.

According to the convex property and optimal solution of each sub-problem, the objective function in Eq.~\eqref{overallobj} will decrease monotonically in each iteration until convergence. For clarity, the procedure of solving the proposed AIMC model is summarized in Algorithm~\ref{Alg:AIalg}.

\begin{table}[!t]
	\caption{Statistics of the nine real-world datasets.}
	\label{TableDataset}
	\centering
	\begin{tabular}{@{}c@{}ccc@{}c@{}}
		\hline     Datasets    & \#Object  & \#View           & \#Class  & View dimension     \\
		\hline HandWritten            & 2000             & 6             & 10       & 216, 76, 64, 6, 240, 47    \\
		Notting-Hill            & 4660               &  3             & 5         & 6750, 3304, 2000        \\
		SUNRGBD            & 10335               &  2             & 45         & 4096, 4096        \\
		AwA            & 30475               &  6             & 50         &  2688, 2000, 252, 2000, 2000, 2000        \\
		VGGFace2-50          & 34027            & 4              & 50       & 944, 576, 512, 640        \\
		YTF-10            & 38654               & 4             & 10         & 944, 576, 512, 640         \\
		YTF-20                & 63896           & 4   &20    & 944, 576, 512, 640   \\
		YTF-50         & 126054              & 4                 & 50         & 944, 576, 512, 640        \\
		YTF-100         & 195537              & 4                 & 100         & 944, 576, 512, 640         \\
		\hline
	\end{tabular}
\end{table}

\textbf{Remark.} In particular, the view generation model in the proposed AIMC method is exploited to discover the underlying latent integral space, which reflects the data structure as well as cluster partition in essence, and avoids the curse of dimensionality to some degree. To validate the effectiveness of AIMC, we additionally employ a naive orthogonal and non-negative matrix factorization method for multiview clustering (NONMF), whose formulation can be rewritten as
\begin{equation}
	\begin{aligned}
		\label{naiveobj}
		&\mathop{\min}\limits_{F^{(v)},Y}\sum_{v=1}^{m}\|X^{(v)}-F^{(v)}Y\|_F^2,\\
		&\text{s.t.~~~}{F^{(v)}}^TF^{(v)}=I_k,Y\in Ind,
	\end{aligned}
\end{equation}
where $F^{(v)}$ is the view-specific cluster centroid matrix with orthogonality constraint, and $Y$ denotes the shared cluster assignment representation. Due to the direct factorization of original data matrix, the effectiveness and efficiency of NONMF would decrease significantly when the data feature dimension is large. Besides, NONMF can not seek for the underlying data space, failing to take the latent semantics among views into account. More details can be found in Section~\ref{sec:experiments}.

\subsection{Complexity Analysis}

\textbf{Computational complexity.} The computational burden of AIMC is composed of four parts, i.e., the cost of optimization about each variable. The first part is to update $G^{(v)}$, whose cost is $O\left(d_vd^2\right)$ and $O\left(d_vdk^2\right)$ respectively for SVD and matrix multiplication. The second part is about the $F$ subproblem, which requires $O\left(dk^2\right)$ for performing SVD on $J$ and $O\left(dk^3\right)$ for matrix multiplication similar to $G^{(v)}$. The third part is to compute the final consensus cluster assignment matrix $Y$, and $O\left(mnk\right)$ is needed. The last part is to solve the weight $\alpha_v$, and it only costs $O(1)$. Therefore, the overall computational complexity to solve the proposed AIMC method is $O\left((hd^2+hdk^2+dk^2+dk^3+mnk)t\right)$, in which $h=\sum_{v=1}^{m}d_v$ is the summation of feature dimensions, and $t$ stands for the number of iterations of these four parts. Consequently, due to the property of large scale datasets (i.e., $n\gg d, k, m$), the computational complexity of AIMC is nearly linear to the number of samples $O\left(n\right)$.

\textbf{Space complexity.} The major space usages of the proposed AIMC method are matrices $G^{(v)}\in\mathbb{R}^{d_v\times d}$, $F\in\mathbb{R}^{d\times k}$, and $Y\in\mathbb{R}^{k\times n}$. Consequently, the major space complexity of AIMC is $O\big(nk+(h+k)d\big)$, which is nearly linear to the number of samples $O\left(n\right)$ as well.

\section{Experiments}
\label{sec:experiments}

\begin{table}[!t]
	\caption{Complexity analysis on compared methods and AIMC. ``Max Reported'' represents the largest dataset size reported in the original reference.}
	\label{TableMemory}
	\centering
	\renewcommand{\arraystretch}{1.2}
	\begin{tabular}{@{}cc@{}cc@{}}
		\hline     Method    & Space Complexity  & Time Complexity          & Max Reported     \\
		\hline AMGL            & $O\left(mn^2+nk\right)$           & $O\left(n^3\right)$               & 12643       \\
		SwMC            & $O\left(n^2+nk\right)$            & $O\left(kn^2\right)$             & 2000    \\
		GMC            & $O\left((m+1)n^2+nk\right)$               &  $O\left((m+k)n^2\right)$             & 11025         \\
		PMSC         & $O\left(2mn^2+(m+1)nk\right)$            & $O\left(n^3\right)$              & 2386            \\
		\hline
		LMVSC            & $O\left(mk(n+h)\right)$               & $O\left(n\right)$             & 30,000                \\
		SMVSC               & $O\left(mn+(h+m)k\right)$           & $O\left(n\right)$   &101,499       \\
		FPMVS         & $O\left(nk+(h+k)k\right)$              & $O\left(n\right)$                 & 101,499              \\
		OPMC         & $O\left(nk+(h+k)k\right)$              & $O\left(n\right)$                 & 101,499              \\
		NONMF           & $O\left(nk+hd\right)$                                      & $O\left(n\right)$            & $\llap{|}$ \\
		AIMC         & $O\left(nk+(h+k)d\right)$              & $O\left(n\right)$                 & 195,537                 \\
		\hline
	\end{tabular}
\end{table}

In this section, to verify the effectiveness of the proposed AIMC method, extensive experiments are conducted on nine real-world datasets, including several large-scale multiview image datasets, to compare the proposed AIMC method with four state-of-the-art multiview clustering methods and five large-scale oriented algorithms in terms of clustering performance and running time. In the meantime, parameter analysis and empirical convergence analysis are also conducted for a comprehensive study. All of the experiments are implemented in Matlab 2021a on a standard Window PC with an Intel 2.4-GHz CPU and 64-GB RAM (64-bit).

\begin{table*}[!t]
	\centering
	\caption{Comparison results in terms of ACC on nine real-world datasets. N/A implies the out-of-memory error.}
	\label{table:ComparisonResultsI}
	\begin{tabular}{m{0.75cm}<{\centering}m{1.45cm}<{\centering}m{1.45cm}<{\centering}m{1.45cm}<{\centering}m{1.45cm}<{\centering}m{1.65cm}<{\centering}m{1.45cm}<{\centering}m{1.45cm}<{\centering}m{1.45cm}<{\centering}m{1.45cm}}
		\toprule
		\multirow{1}{*}{Methods} &\multicolumn{1}{c}{HandWritten} &\multicolumn{1}{c}{Notting-Hill} &\multicolumn{1}{c}{SUNRGBD} &\multicolumn{1}{c}{AwA} &\multicolumn{1}{c}{VGGFace2-50} &\multicolumn{1}{c}{YTF-10} &\multicolumn{1}{c}{YTF-20} &\multicolumn{1}{c}{YTF-50} &\multicolumn{1}{c}{YTF-100} \\
		\hline
		\multirow{1}*{AMGL}
		&0.8082$_{\pm0.0566}$	&0.3609$_{\pm0.0178}$	&0.2143$_{\pm0.0119}$	&N/A	&N/A	&N/A	&N/A   &N/A	   &N/A \\
		\multirow{1}*{SwMC}
		&0.8740$_{\pm0.0000}$	&0.3388$_{\pm0.0000}$	&0.1181$_{\pm0.0000}$	&N/A	&N/A	&N/A	&N/A   &N/A	   &N/A \\
		\multirow{1}*{GMC}
		&0.8820$_{\pm0.0891}$	&0.3124$_{\pm0.0000}$	&0.1277$_{\pm0.0000}$	&N/A	&N/A	&N/A	&N/A   &N/A	   &N/A \\
		\multirow{1}*{PMSC}
		&0.6318$_{\pm0.0440}$	&0.7782$_{\pm0.0617}$	&N/A	&N/A	&N/A	&N/A	&N/A   &N/A	   &N/A \\
		\multirow{1}*{LMVSC}
		&\underline{0.9225$_{\pm0.0000}$} &\textbf{0.9133$_{\pm0.0000}$} &0.1810$_{\pm0.0000}$ &0.0760$_{\pm0.0000}$ &0.1113$_{\pm0.0000}$ &0.7566$_{\pm0.0000}$ &0.7092$_{\pm0.0000}$ &0.6825$_{\pm0.0000}$ &0.6006$_{\pm0.0000}$ \\
		\multirow{1}*{SMVSC}
		&0.8205$_{\pm0.0000}$ &0.8081$_{\pm0.0000}$ &0.2223$_{\pm0.0000}$ &0.0908$_{\pm0.0000}$ &0.1127$_{\pm0.0000}$ &0.7389$_{\pm0.0000}$ &\underline{0.7125$_{\pm0.0000}$} &0.6681$_{\pm0.0000}$ &0.5907$_{\pm0.0000}$ \\
		\multirow{1}*{FPMVS}
		&0.8230$_{\pm0.0000}$ &0.7131$_{\pm0.0000}$ &\underline{0.2388$_{\pm0.0000}$} &0.0912$_{\pm0.0000}$ &0.1098$_{\pm0.0000}$ &0.7325$_{\pm0.0000}$ &0.6948$_{\pm0.0000}$ &0.6851$_{\pm0.0000}$ &0.5293$_{\pm0.0000}$\\
		\multirow{1}*{OPMC}
		&0.8915$_{\pm0.0000}$ &0.7431$_{\pm0.0000}$ &0.1930$_{\pm0.0000}$ &0.0925$_{\pm0.0000}$ &\underline{0.1160$_{\pm0.0000}$} &0.7532$_{\pm0.0000}$ &0.6782$_{\pm0.0000}$ &\underline{0.7165$_{\pm0.0000}$} &0.6443$_{\pm0.0000}$\\
		\multirow{1}*{NONMF}
		&0.8400$_{\pm0.0000}$ &0.8027$_{\pm0.0000}$ &0.2324$_{\pm0.0000}$ &\underline{0.0965$_{\pm0.0000}$} &0.1054$_{\pm0.0000}$ &\underline{0.7726$_{\pm0.0000}$} &0.6509$_{\pm0.0000}$ &0.7135$_{\pm0.0000}$ &\underline{0.6610$_{\pm0.0000}$}\\
		\multirow{1}*{AIMC}
		&\textbf{0.9345$_{\pm0.0000}$} &\underline{0.9064$_{\pm0.0000}$} &\textbf{0.2536$_{\pm0.0000}$} &\textbf{0.1008$_{\pm0.0000}$} &\textbf{0.1186$_{\pm0.0000}$} &\textbf{0.8881$_{\pm0.0000}$} &\textbf{0.7640$_{\pm0.0000}$} &\textbf{0.7602$_{\pm0.0000}$} &\textbf{0.6789$_{\pm0.0000}$}\\
		\bottomrule
	\end{tabular}
\end{table*}

\begin{table*}[!t]
	\centering
	\caption{Comparison results in terms of NMI on nine real-world datasets. N/A implies the out-of-memory error.}
	\label{table:ComparisonResultsII}
	\begin{tabular}{m{0.75cm}<{\centering}m{1.45cm}<{\centering}m{1.45cm}<{\centering}m{1.45cm}<{\centering}m{1.45cm}<{\centering}m{1.65cm}<{\centering}m{1.45cm}<{\centering}m{1.45cm}<{\centering}m{1.45cm}<{\centering}m{1.45cm}}
		\toprule
		\multirow{1}{*}{Methods} &\multicolumn{1}{c}{HandWritten} &\multicolumn{1}{c}{Notting-Hill} &\multicolumn{1}{c}{SUNRGBD} &\multicolumn{1}{c}{AwA} &\multicolumn{1}{c}{VGGFace2-50} &\multicolumn{1}{c}{YTF-10} &\multicolumn{1}{c}{YTF-20} &\multicolumn{1}{c}{YTF-50} &\multicolumn{1}{c}{YTF-100} \\
		\hline
		\multirow{1}*{AMGL}
		&0.8463$_{\pm0.0294}$	&0.0780$_{\pm0.0208}$	&0.1686$_{\pm0.0109}$    &N/A	&N/A	&N/A	&N/A   &N/A	   &N/A \\
		\multirow{1}*{SwMC}
		&\underline{0.8886$_{\pm0.0000}$}	&0.0877$_{\pm0.0000}$	&0.0287$_{\pm0.0000}$	&N/A	&N/A	&N/A	&N/A   &N/A	   &N/A \\
		\multirow{1}*{GMC}
		&\textbf{0.9041$_{\pm0.0000}$}	&0.0923$_{\pm0.0000}$	&0.0728$_{\pm0.0000}$	&N/A	&N/A	&N/A	&N/A   &N/A	   &N/A \\
		\multirow{1}*{PMSC}
		&0.6567$_{\pm0.0391}$	&0.6563$_{\pm0.0430}$	&N/A	&N/A	&N/A	&N/A   &N/A	   &N/A    &N/A \\
		\multirow{1}*{LMVSC}
		&0.8576$_{\pm0.0000}$ &\underline{0.7966$_{\pm0.0000}$} &0.2283$_{\pm0.0000}$	&0.0879$_{\pm0.0000}$ &0.1332$_{\pm0.0000}$ &0.7670$_{\pm0.0000}$ &0.7751$_{\pm0.0000}$ &0.8098$_{\pm0.0000}$ &0.7804$_{\pm0.0000}$ \\
		\multirow{1}*{SMVSC}
		&0.7886$_{\pm0.0000}$ &0.7371$_{\pm0.0000}$ &0.2280$_{\pm0.0000}$	&0.1089$_{\pm0.0000}$ &\underline{0.1366$_{\pm0.0000}$} &0.7980$_{\pm0.0000}$ &0.7913$_{\pm0.0000}$ &0.8258$_{\pm0.0000}$ &0.7991$_{\pm0.0000}$ \\
		\multirow{1}*{FPMVS}
		&0.7923$_{\pm0.0000}$ &0.6772$_{\pm0.0000}$ &0.2413$_{\pm0.0000}$	&0.1088$_{\pm0.0000}$ &0.1365$_{\pm0.0000}$ &0.7740$_{\pm0.0000}$ &0.7790$_{\pm0.0000}$ &0.8364$_{\pm0.0000}$ &0.7532$_{\pm0.0000}$\\
		\multirow{1}*{OPMC}
		&0.8131$_{\pm0.0000}$ &0.7638$_{\pm0.0000}$ &\textbf{0.2523$_{\pm0.0000}$}	&\underline{0.1195$_{\pm0.0000}$} &0.1344$_{\pm0.0000}$ &0.8014$_{\pm0.0000}$ &0.7726$_{\pm0.0000}$ &0.8278$_{\pm0.0000}$ &0.8079$_{\pm0.0000}$\\
		\multirow{1}*{NONMF}
		&0.7981$_{\pm0.0000}$ &0.7126$_{\pm0.0000}$ &0.2359$_{\pm0.0000}$	&0.1173$_{\pm0.0000}$ &0.1324$_{\pm0.0000}$ &\underline{0.8144$_{\pm0.0000}$} &\underline{0.7952$_{\pm0.0000}$} &\underline{0.8479$_{\pm0.0000}$} &\underline{0.8280$_{\pm0.0000}$}\\
		\multirow{1}*{AIMC}
		&0.8823$_{\pm0.0000}$ &\textbf{0.8226$_{\pm0.0000}$} &\underline{0.2501$_{\pm0.0000}$}	&\textbf{0.1200$_{\pm0.0000}$} &\textbf{0.1489$_{\pm0.0000}$} &\textbf{0.8704$_{\pm0.0000}$} &\textbf{0.8334$_{\pm0.0000}$} &\textbf{0.8643$_{\pm0.0000}$} &\textbf{0.8431$_{\pm0.0000}$}\\
		\bottomrule
	\end{tabular}
\end{table*}

\begin{table*}[!t]
	\centering
	\caption{Comparison results in terms of Purity on nine real-world datasets. N/A implies the out-of-memory error.}
	\label{table:ComparisonResultsIII}
	\begin{tabular}{m{0.75cm}<{\centering}m{1.45cm}<{\centering}m{1.45cm}<{\centering}m{1.45cm}<{\centering}m{1.45cm}<{\centering}m{1.65cm}<{\centering}m{1.45cm}<{\centering}m{1.45cm}<{\centering}m{1.45cm}<{\centering}m{1.45cm}}
		\toprule
		\multirow{1}{*}{Methods} &\multicolumn{1}{c}{HandWritten} &\multicolumn{1}{c}{Notting-Hill} &\multicolumn{1}{c}{SUNRGBD} &\multicolumn{1}{c}{AwA} &\multicolumn{1}{c}{VGGFace2-50} &\multicolumn{1}{c}{YTF-10} &\multicolumn{1}{c}{YTF-20} &\multicolumn{1}{c}{YTF-50} &\multicolumn{1}{c}{YTF-100} \\
		\hline
		\multirow{1}*{AMGL}
		&0.8281$_{\pm0.0475}$	&0.3677$_{\pm0.0161}$	&0.2631$_{\pm0.0087}$    &N/A	&N/A	&N/A	&N/A   &N/A	   &N/A \\
		\multirow{1}*{SwMC}
		&0.8795$_{\pm0.0000}$	&0.3710$_{\pm0.0000}$	&0.1311$_{\pm0.0000}$    &N/A	&N/A	&N/A	&N/A   &N/A	   &N/A \\
		\multirow{1}*{GMC}
		&0.8820$_{\pm0.0000}$	&0.3380$_{\pm0.0000}$	&0.1415$_{\pm0.0000}$    &N/A	&N/A	&N/A	&N/A   &N/A	   &N/A \\
		\multirow{1}*{PMSC}
		&0.8311$_{\pm0.0363}$	&0.7948$_{\pm0.0488}$    &N/A	&N/A	&N/A	&N/A   &N/A	   &N/A  &N/A \\
		\multirow{1}*{LMVSC}
		&0.9225$_{\pm0.0000}$ &\textbf{0.9133$_{\pm0.0000}$} &0.1856$_{\pm0.0000}$	&0.0868$_{\pm0.0000}$ &\textbf{0.1401$_{\pm0.0000}$} &\underline{0.8125$_{\pm0.0000}$} &0.7614$_{\pm0.0000}$ &\underline{0.7688$_{\pm0.0000}$} &0.6827$_{\pm0.0000}$ \\
		\multirow{1}*{SMVSC}
		&0.8205$_{\pm0.0000}$ &0.8324$_{\pm0.0000}$ &0.3292$_{\pm0.0000}$	&0.0966$_{\pm0.0000}$ &0.1157$_{\pm0.0000}$ &0.7726$_{\pm0.0000}$ &0.7700$_{\pm0.0000}$ &0.6926$_{\pm0.0000}$ &0.6104$_{\pm0.0000}$ \\
		\multirow{1}*{FPMVS}
		&0.8230$_{\pm0.0000}$ &0.7850$_{\pm0.0000}$ &0.3399$_{\pm0.0000}$	&0.0963$_{\pm0.0000}$ &0.1142$_{\pm0.0000}$ &0.7621$_{\pm0.0000}$ &0.7259$_{\pm0.0000}$ &0.7140$_{\pm0.0000}$ &0.5446$_{\pm0.0000}$\\
		\multirow{1}*{OPMC}
		&0.8915$_{\pm0.0000}$ &0.8285$_{\pm0.0000}$ &\textbf{0.3974$_{\pm0.0000}$}	&\textbf{0.1140$_{\pm0.0000}$} &0.1241$_{\pm0.0000}$ &0.8061$_{\pm0.0000}$ &0.7250$_{\pm0.0000}$ &0.7686$_{\pm0.0000}$ &0.6983$_{\pm0.0000}$\\
		\multirow{1}*{NONMF}
		&0.8400$_{\pm0.0000}$ &0.8027$_{\pm0.0000}$ &0.3327$_{\pm0.0000}$	&0.1025$_{\pm0.0000}$ &\underline{0.1324$_{\pm0.0000}$} &0.8057$_{\pm0.0000}$ &0.7381$_{\pm0.0000}$ &0.7531$_{\pm0.0000}$ &\underline{0.6995$_{\pm0.0000}$}\\
		\multirow{1}*{AIMC}
		&\textbf{0.9345$_{\pm0.0000}$} &\underline{0.9064$_{\pm0.0000}$} &\underline{0.3519$_{\pm0.0000}$}	&\underline{0.1097$_{\pm0.0000}$} &0.1271$_{\pm0.0000}$ &\textbf{0.8881$_{\pm0.0000}$} &\textbf{0.8110$_{\pm0.0000}$} &\textbf{0.7879$_{\pm0.0000}$} &\textbf{0.7267$_{\pm0.0000}$}\\
		\bottomrule
	\end{tabular}
\end{table*}

\begin{table*}[!t]
	\centering
	\caption{Comparison results in terms of Fscore on nine real-world datasets. N/A implies the out-of-memory error.}
	\label{table:ComparisonResultsIV}
	\begin{tabular}{m{0.75cm}<{\centering}m{1.45cm}<{\centering}m{1.45cm}<{\centering}m{1.45cm}<{\centering}m{1.45cm}<{\centering}m{1.65cm}<{\centering}m{1.45cm}<{\centering}m{1.45cm}<{\centering}m{1.45cm}<{\centering}m{1.45cm}}
		\toprule
		\multirow{1}{*}{Methods} &\multicolumn{1}{c}{HandWritten} &\multicolumn{1}{c}{Notting-Hill} &\multicolumn{1}{c}{SUNRGBD} &\multicolumn{1}{c}{AwA} &\multicolumn{1}{c}{VGGFace2-50} &\multicolumn{1}{c}{YTF-10} &\multicolumn{1}{c}{YTF-20} &\multicolumn{1}{c}{YTF-50} &\multicolumn{1}{c}{YTF-100} \\
		\hline
		\multirow{1}*{AMGL}
		&0.7681$_{\pm0.0853}$	&0.3654$_{\pm0.0040}$	&0.1422$_{\pm0.0046}$    &N/A	&N/A	&N/A	&N/A   &N/A	   &N/A \\
		\multirow{1}*{SwMC}
		&0.8624$_{\pm0.0000}$	&0.3690$_{\pm0.0000}$	&0.1204$_{\pm0.0000}$    &N/A	&N/A	&N/A	&N/A   &N/A	   &N/A \\
		\multirow{1}*{GMC}
		&\underline{0.8653$_{\pm0.0000}$}	&0.3694$_{\pm0.0000}$	&0.1215$_{\pm0.0000}$    &N/A	&N/A	&N/A	&N/A   &N/A	   &N/A \\
		\multirow{1}*{PMSC}
		&0.5539$_{\pm0.0476}$	&0.6568$_{\pm0.0565}$    &N/A	&N/A	&N/A	&N/A   &N/A	   &N/A   &N/A \\
		\multirow{1}*{LMVSC}
		&0.8499$_{\pm0.0000}$ &\underline{0.8347$_{\pm0.0000}$} &0.1154$_{\pm0.0000}$	&0.0377$_{\pm0.0000}$ &0.0524$_{\pm0.0000}$ &0.7001$_{\pm0.0000}$ &0.6268$_{\pm0.0000}$ &0.5794$_{\pm0.0000}$ &0.5171$_{\pm0.0000}$ \\
		\multirow{1}*{SMVSC}
		&0.7525$_{\pm0.0000}$ &0.7870$_{\pm0.0000}$ &0.1487$_{\pm0.0000}$	&\textbf{0.0641$_{\pm0.0000}$} &\underline{0.0604$_{\pm0.0000}$} &0.6982$_{\pm0.0000}$ &0.6518$_{\pm0.0000}$ &0.6157$_{\pm0.0000}$ &0.5036$_{\pm0.0000}$ \\
		\multirow{1}*{FPMVS}
		&0.7556$_{\pm0.0000}$ &0.6928$_{\pm0.0000}$ &\underline{0.1604$_{\pm0.0000}$}	&\underline{0.0640$_{\pm0.0000}$} &0.0597$_{\pm0.0000}$ &0.6959$_{\pm0.0000}$ &0.6261$_{\pm0.0000}$ &0.6381$_{\pm0.0000}$ &0.3541$_{\pm0.0000}$\\
		\multirow{1}*{OPMC}
		&0.8033$_{\pm0.0000}$ &0.7395$_{\pm0.0000}$ &0.1339$_{\pm0.0000}$	&0.0469$_{\pm0.0000}$ &0.0550$_{\pm0.0000}$ &0.7225$_{\pm0.0000}$ &0.6206$_{\pm0.0000}$ &0.6376$_{\pm0.0000}$ &\underline{0.5651$_{\pm0.0000}$}\\
		\multirow{1}*{NONMF}
		&0.7757$_{\pm0.0000}$ &0.7802$_{\pm0.0000}$ &0.1506$_{\pm0.0000}$	&0.0634$_{\pm0.0000}$ &0.0580$_{\pm0.0000}$ &\underline{0.7461$_{\pm0.0000}$} &\underline{0.6590$_{\pm0.0000}$} &\underline{0.6573$_{\pm0.0000}$} &0.5463$_{\pm0.0000}$\\
		\multirow{1}*{AIMC}
		&\textbf{0.8790$_{\pm0.0000}$} &\textbf{0.8565$_{\pm0.0000}$} &\textbf{0.1709$_{\pm0.0000}$}	&\underline{0.0640$_{\pm0.0000}$} &\textbf{0.0606$_{\pm0.0000}$} &\textbf{0.8288$_{\pm0.0000}$} &\textbf{0.7372$_{\pm0.0000}$} &\textbf{0.7004$_{\pm0.0000}$} &\textbf{0.6161$_{\pm0.0000}$}\\
		\bottomrule
	\end{tabular}
\end{table*}

\begin{table*}[!t]
	\caption{Comparison results on nine real-world datasets in terms of running time in seconds. N/A implies the out-of-memory error.}
	\label{table:runtime}
	\centering
	\renewcommand{\arraystretch}{1.3}
	\begin{tabular}{ccccccccccc}
		\hline   Datasets~ &~AMGL~ &~SwMC~ &~GMC~ &~PMSC~ & ~LMVSC~ & ~SMVSC~ & ~FPMVS~ & ~OPMC~ & ~NONMF~ & ~AIMC~ \\
		\hline HandWritten  &40.65~ &144.71~&21.75~&2133.26~&\underline{2.20}~&14.58~&21.49~&\textbf{0.97}~&6.13~&2.60~ \\
		Notting-Hill  &592.33~ &3276.04~ &27.72~ &3287.20~ &\textbf{2.35}~ &375.08~ &729.19~ &\underline{20.21}~ &34.45~ &21.57~ \\
		SUNRGBD  &4504.63~ &1522.83~ &483.24~ &N/A~ &\textbf{8.94}~ &551.70~ &1070.34~ &145.45~ &122.22~ &\underline{51.42}~ \\
		AwA  &N/A~ &N/A~ &N/A~ &N/A~ &\textbf{55.14}~ &2265.28~ &4025.50~ &897.73~ &301.95~ &\underline{166.15}~ \\
		VGGFace2-50~	  &N/A~ &N/A~ &N/A~ &N/A~ &\textbf{21.98}~ &820.42~ &1279.40~ &256.09~ &131.42~ &\underline{76.16}~ \\
		YTF-10~	 &N/A~ &N/A~ &N/A~ &N/A~ &\textbf{6.78}~ &903.47~ &1378.40~ &\underline{47.85}~ &60.55~ &49.91~  \\
		YTF-20~  &N/A~ &N/A~ &N/A~ &N/A~ &\textbf{15.34}~ &1906.92~ &2717.5~ &\underline{60.33}~ &153.77~ &76.84~ \\
		YTF-50~  &N/A~ &N/A~ &N/A~ &N/A~ &\textbf{119.02}~ &3981.79~ &7030.41~ &245.94~ &284.69~ &\underline{180.49}~ \\
		YTF-100~ &N/A~ &N/A~ &N/A~ &N/A~ &\underline{345.61}~ &16331.04~ &19978.28~ &576.28~ &404.28~ &\textbf{336.32}~ \\
		\hline
		Rank~   &8.0~      &8.7~      &5.3~     &10.0~     &\textbf{1.2}~     &5.2~      &6.3~     &2.9~     &3.7~     &\underline{2.3}~ \\
		\hline
	\end{tabular}
\end{table*}

\subsection{Datasets and Compared Methods}

Nine widely used datasets are adopted in the evaluation, whose statistics are summarized in Table~\ref{TableDataset}. \textbf{HandWritten} consists of ten-class handwritten digits, and each digit has 200 samples~\cite{DBLP:journals/kybernetika/BreukelenDTH98}. In our experiments, six features are selected, which are 216-D profile correlation, 76-D Fourier coefficient, 64-D Karhunen-Love coefficient, 6-D morphological feature, 240-D pixel average and 47-D Zernike moment. \textbf{Notting-Hill} contains 4660 facial images where the faces of five main casts are selected~\cite{zhang2009character}. In our experiments, three features are selected, including 6750-D Gabor feature, 3304-D LBP feature and 2000-D intensity feature. \textbf{SUNRGBD} consists of 10335 indoor scene images of 45 object categories collected from Princeton University~\cite{DBLP:conf/cvpr/SongLX15}. In our experiments, two features are selected, whose dimensions are both 4096. \textbf{AwA} is an animal dataset consisting of 30475 images belonging to 50 animal categories~\cite{DBLP:journals/pami/LampertNH14}. In our experiments, six features are extracted, which are 2688-D color histogram, 2000-D local self-similarity, 252-D PHOG feature, 2000-D SIFT feature, 2000-D color SIFT feature and 2000-D SURF feature. \textbf{VGGFace2-50}\footnotemark[1] is a face image dataset containing 34027 images in 50 classes. In our experiments, four features are selected, which are 944-D LBP feature, 576-D HOG feature, 512-D GIST feature and 640-D Gabor feature. \textbf{YTF-10} (YouTube Faces), \textbf{YTF-20}, \textbf{YTF-50} and \textbf{YTF-100} are different subsets of face videos obtained from YouTube\footnotemark[2], whose data size are respectively 38654, 63896, 126054 and 195537. In our experiments, features of four views are extracted in all these four datasets, including 944-D LBP feature, 576-D HOG feature, 512-D GIST feature and 640-D Gabor feature.
\footnotetext[1]{\url{https://www.robots.ox.ac.uk/~vgg/data/vgg_face2/.}}
\footnotetext[2]{\url{https://www.cs.tau.ac.il/~wolf/ytfaces/.}}

The proposed AIMC method is compared with the following state-of-the-art baselines. \textbf{Parameter-free Auto-weighted Multiple Graph Learning (AMGL)}~\cite{DBLP:conf/ijcai/NieLL16} proposes to learn the optimal weights for each graph. \textbf{Self-weighted Multiview Clustering with Multiple Graphs (SwMC)}~\cite{DBLP:conf/ijcai/NieLL17} attempts to study a Laplacian rank constrained graph. \textbf{Graph-based Multiview Clustering (GMC)}~\cite{DBLP:journals/tkde/WangYL20} learns similarity-induced graphs from multiple views and unified fusion graph in a mutual manner. \textbf{Partition Level Multiview Subspace Clustering (PMSC)}~\cite{DBLP:journals/nn/KangZPZZPCX20} develops a unified multiview subspace clustering model. \textbf{Large-scale Multiview Subspace Clustering in Linear Time (LMVSC)}~\cite{DBLP:conf/aaai/KangZZSHX20} proposes a multiview subspace clustering framework for large-scale data in which a smaller graph for each view is studied. \textbf{Scalable Multiview Subspace Clustering with Unified Anchors (SMVSC)}~\cite{DBLP:conf/mm/SunZWZTLZW21} attempts to construct graph representation based on actual latent data distribution. \textbf{Fast Parameter-Free Multiview Subspace Clustering With Consensus Anchor Guidance (FPMVS)}~\cite{DBLP:journals/tip/WangLZZZGZ22} studies the subspace representation with consensus anchor guidance. \textbf{One-pass Multiview Clustering for Large-scale Data (OPMC)} considers matrix factorization together with partition generation~\cite{DBLP:conf/iccv/0003LYL0LS21}. \textbf{Naive Orthogonal and Non-negative Matrix Factorization for multiview clustering (NONMF)} makes direct factorization of original data matrix, which is included for ablation study to confirm the superiority of AIMC.

For comparison, we summarize the major space and time complexity of the compared methods in Table~\ref{TableMemory}.

\subsection{Experimental Setup}

For all methods, including the proposed AIMC method and nine compared methods, experiments are conducted twenty times for each of them and the average performance as well as the standard deviation (std. dev.) are reported. In the meantime, $k$-means algorithm is needed to achieve the final clustering performance for most of the methods, and thus in each experiment, it is run ten times to eliminate the random initialization. For our method, the dimension of the view generation model $d$ is tuned from $\left[k:5:300\right]$ for the benchmark datasets, where $d\geq k$ should be satisfied so as to better describe the cluster space. For the compared methods, the best parameters are tuned as suggested by the corresponding papers.

Comprehensively, the clustering performance is evaluated by four widely used metrics, i.e., accuracy (ACC), normalized mutual information (NMI), purity and Fscore. For all the above evaluation metrics, higher values indicate better clustering performance. More detailed descriptions of the four measures can be referred to~\cite{DBLP:journals/tnn/WangCHLY21}. Since each evaluation metric penalizes or favours certain property in clusterings, a more comprehensive evaluation can be received by reporting clustering performance via the four diverse metrics.

\subsection{Comparison Results}
\subsubsection{Clustering Performance}

The clustering performance obtained by different multiview clustering methods on the nine real-world datasets are reported in terms of ACC, NMI, Purity and Fscore in Table~\ref{table:ComparisonResultsI},~\ref{table:ComparisonResultsII},~\ref{table:ComparisonResultsIII} and ~\ref{table:ComparisonResultsIV}, where the mean as well as the standard deviation (std. dev.) are reported over 20 runs. In these tables, the best performance for distinct datasets in terms of each measure is highlighted in boldface, while the second best one in underline. Note that N/A implies that the algorithm suffers out-of-memory error on the corresponding dataset on our device, and thus no result is reported. According to the four tables, the following observations can be obtained.
\begin{enumerate}
	\item In general, the proposed AIMC method achieves the best clustering performance on most of the benchmark datasets, and always obtains better results than other multiview clustering methods for large-scale datasets. For instance, on the VGGFace2-50 dataset, AIMC significantly outperforms the other five large-scale oriented algorithms by achieving improvements of 1.57$\%$, 1.23$\%$, 1.24$\%$, 1.45$\%$ and 1.65$\%$ in terms of NMI. On the YTF-10 dataset, the performance improvements over the other algorithms in terms of NMI are respectively 10.34$\%$, 7.24$\%$, 9.64$\%$, 6.90$\%$ and 5.60$\%$.
	\item It could be concluded that the anchor based methods (LMVSC, SMVSC, FPMVS) or matrix factorization based methods (OPMC, NONMF, AIMC) are more suitable for large-scale datasets while comparing with the traditional multiview subspace clustering methods (AMGL, SwMC, GMC, PMSC).
	\item Despite good results achieved by NONMF, the performance of the proposed AIMC method is obviously superior with meaningful discovery of the latent integral space constructed by data cluster structure.
\end{enumerate}

Consequently, empirical comparison results validate the superiority of the proposed AIMC model.

\subsubsection{Running Time}

For further study, the comparison of running time in seconds consumed by all compared algorithms on the nine real-world datasets is reported in Table~\ref{table:runtime}. From the table, it can be observed that a relatively good trade-off between the clustering performance and computational burden is achieved by the proposed AIMC method, which averagely ranks the second best on computational time. Despite less computational time cost by LMVSC, its clustering performance is worse than AIMC on the whole. This is mainly due to that LMVSC merely focuses on the construction of multiple view-specific graph representations, neglecting the cross-view complementary information and underlying correlations from the latent integral space. Note that on the largest benchmark dataset (i.e., YTF-100), the proposed AIMC method needs the least computational cost, indicating the strong potential for facing with bigger data. As a result, large-scale multiview data clustering problems can be tackled by the proposed AIMC method with efficient computation.

\begin{figure}[!t]
	\centerline{
		\subfigure[HandWritten]{
			\includegraphics[width=0.5\linewidth]{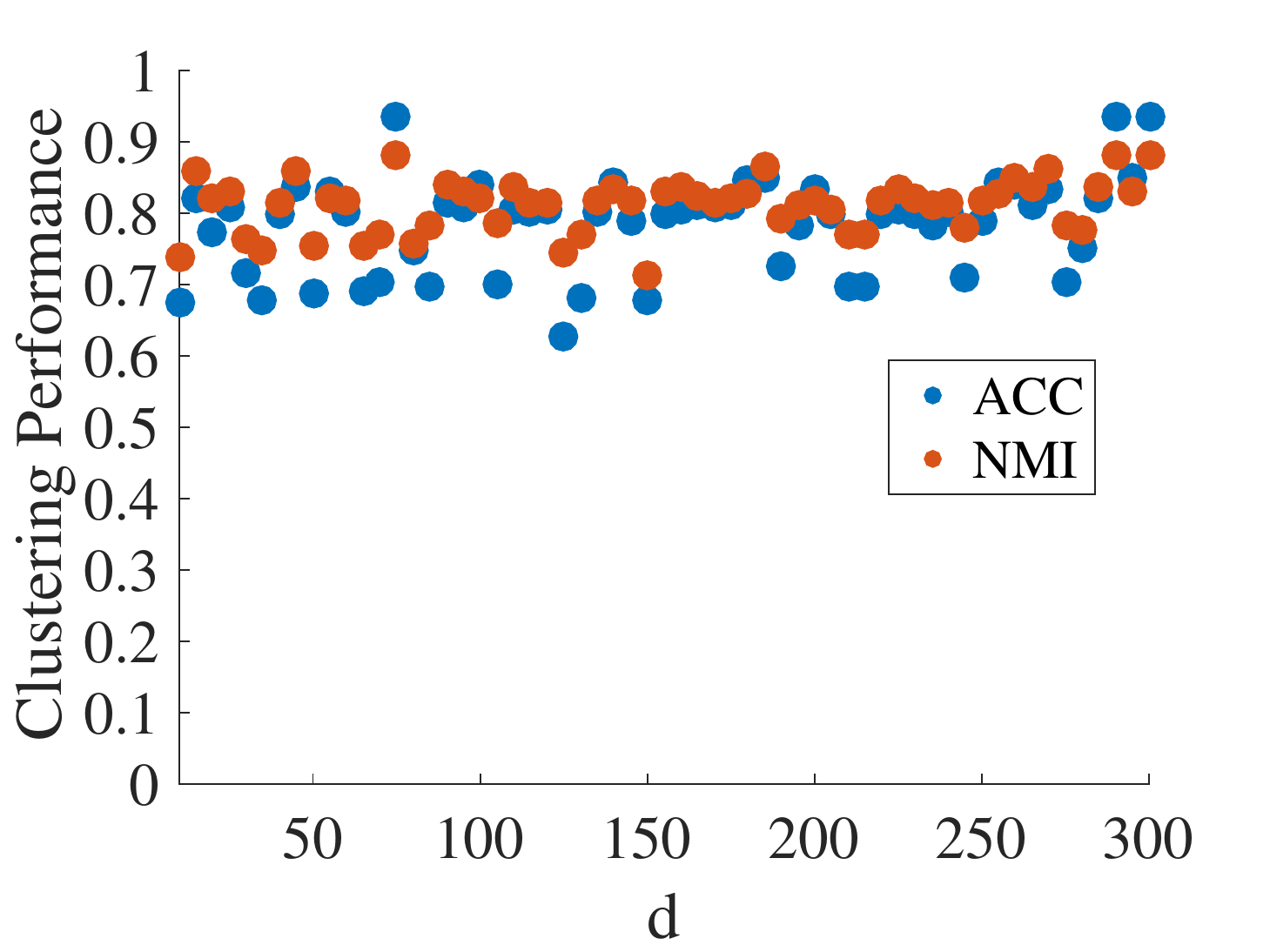}
		}
		\subfigure[SUNRGBD]{
			\includegraphics[width=0.5\linewidth]{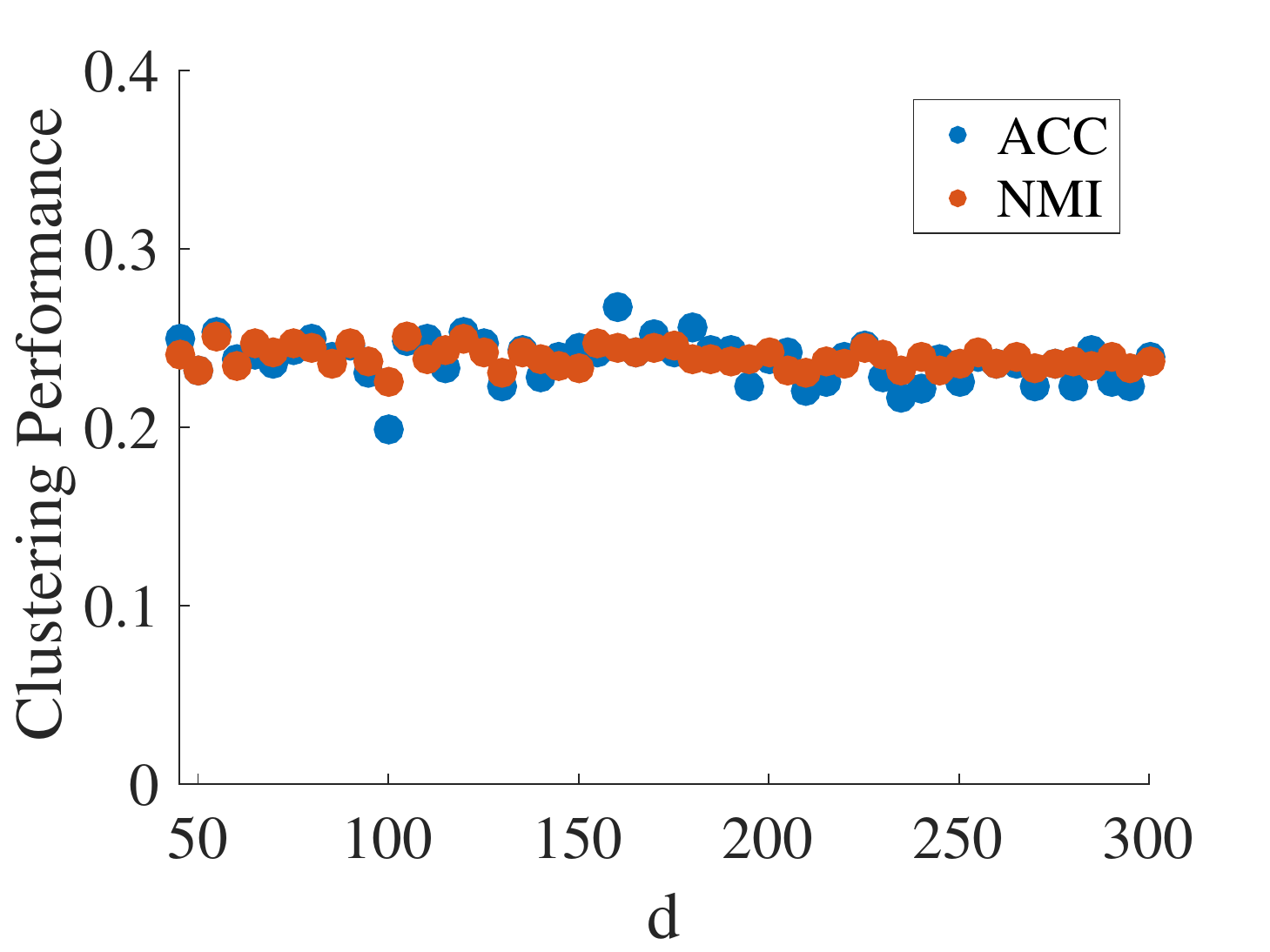}
		}
	}
	\centerline{
		\subfigure[VGGFace2-50]{
			\includegraphics[width=0.5\linewidth]{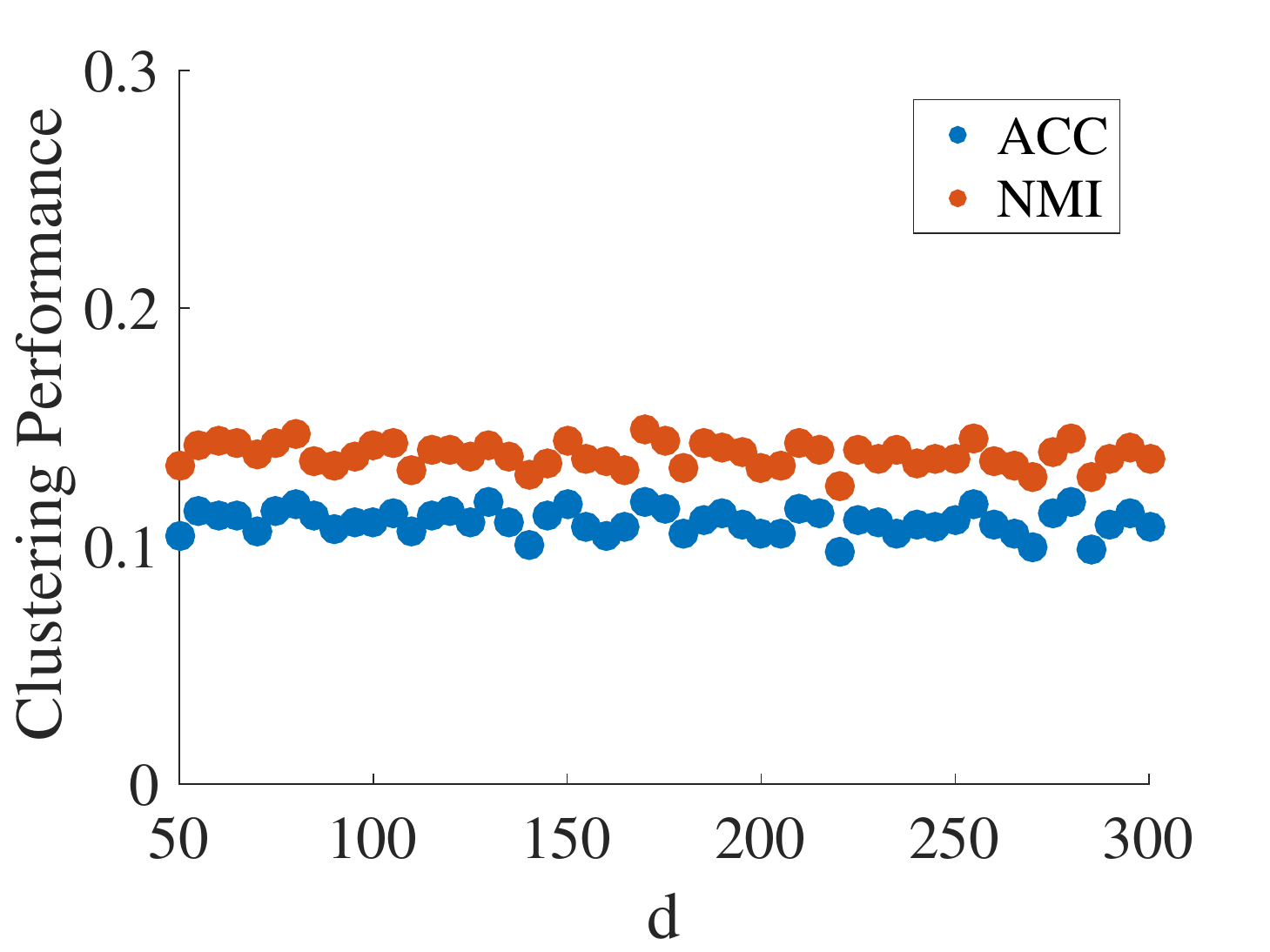}
		}
		\subfigure[YTF-20]{
			\includegraphics[width=0.5\linewidth]{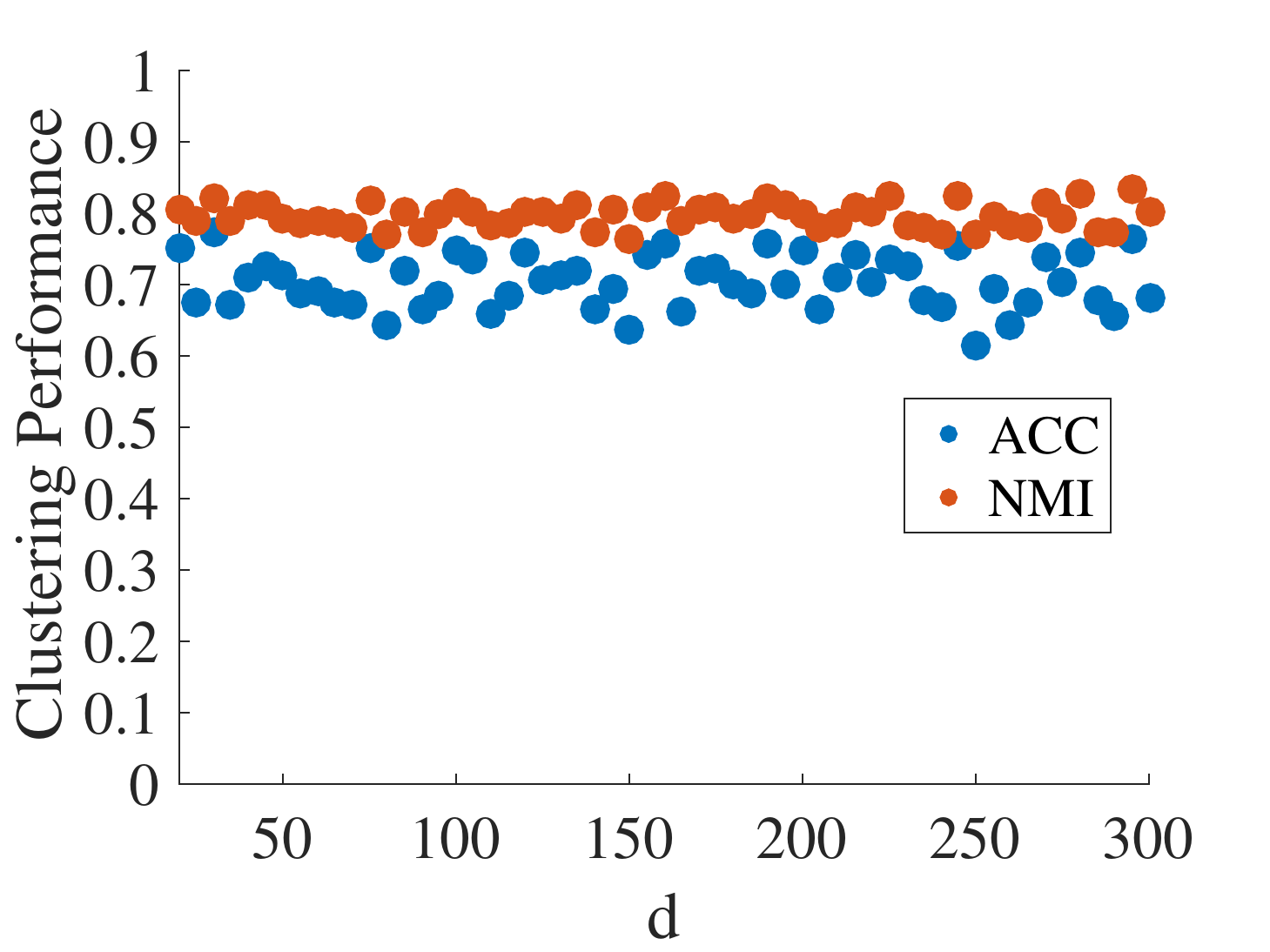}
		}
	}
	\caption{Parameter analysis: Performance in terms of ACC and NMI with different dimension of the view generation model $d$ on the four datasets.}
	\label{Figure:paraAnalysis}
\end{figure}

\begin{figure}[!t]
	\centerline{
		\subfigure[HandWritten]{
			\includegraphics[width=0.5\linewidth]{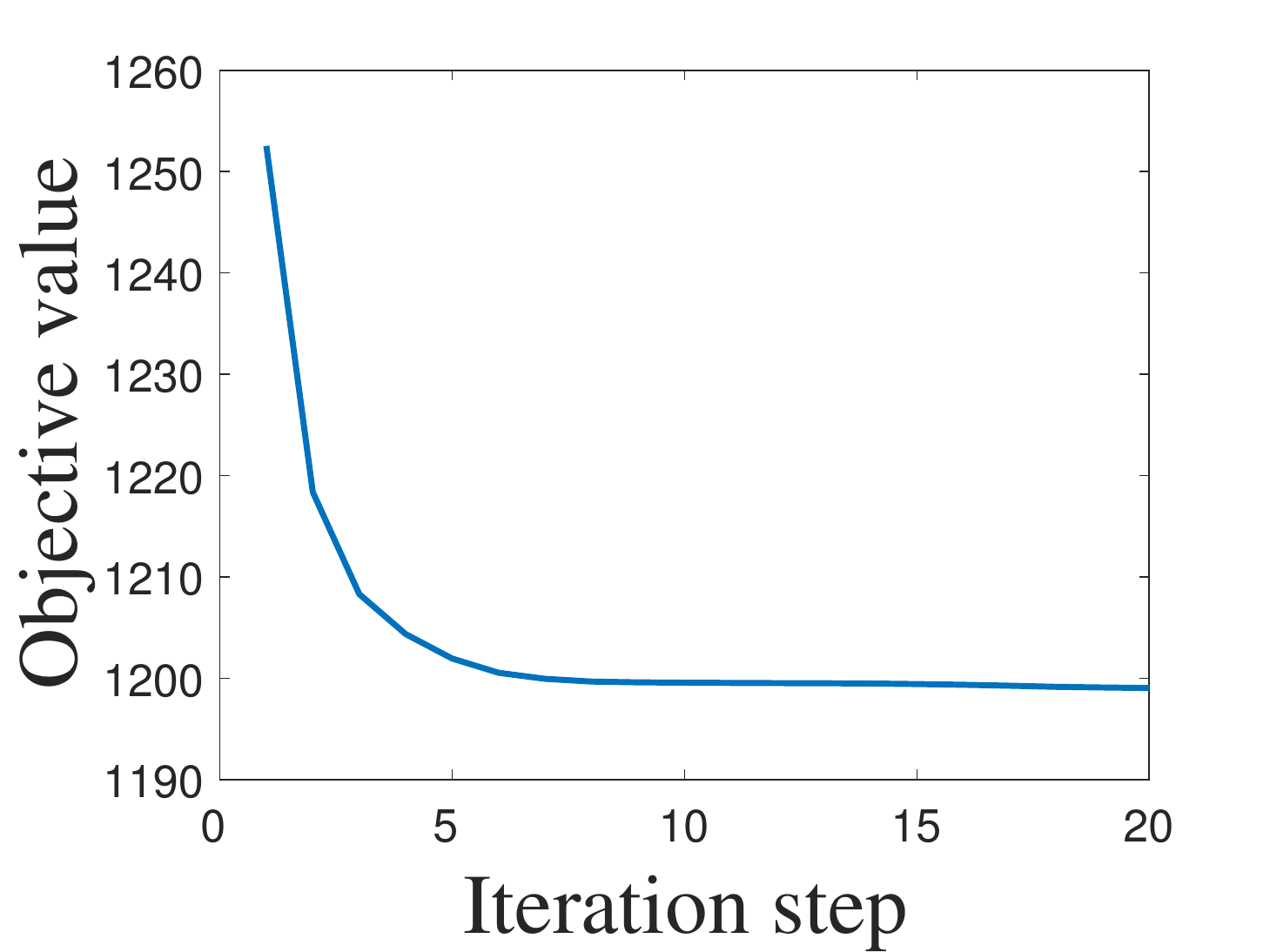}
		}
		\subfigure[SUNRGBD]{
			\includegraphics[width=0.5\linewidth]{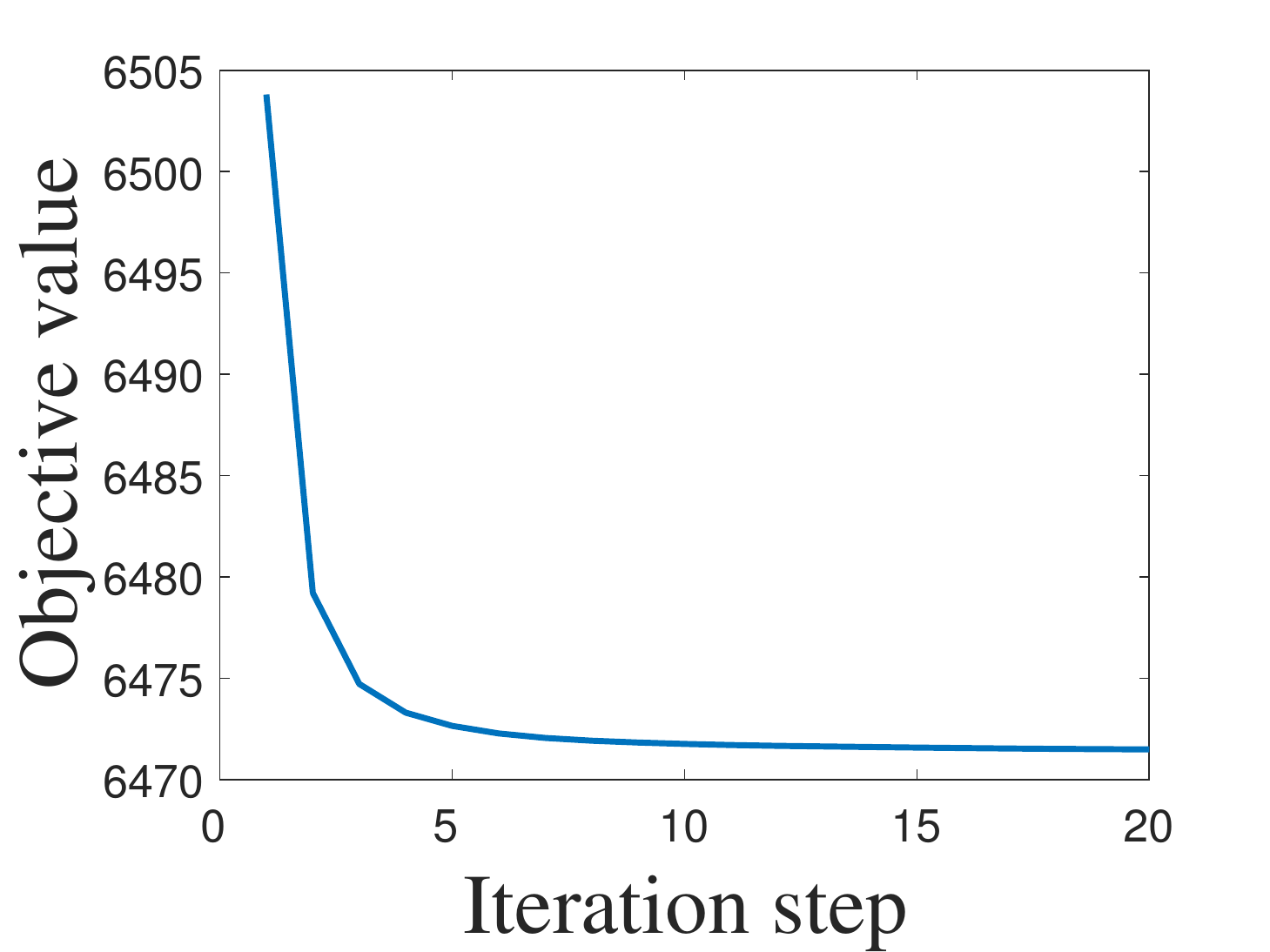}
	}}
	\centerline{
		\subfigure[VGGFace2-50]{
			\includegraphics[width=0.5\linewidth]{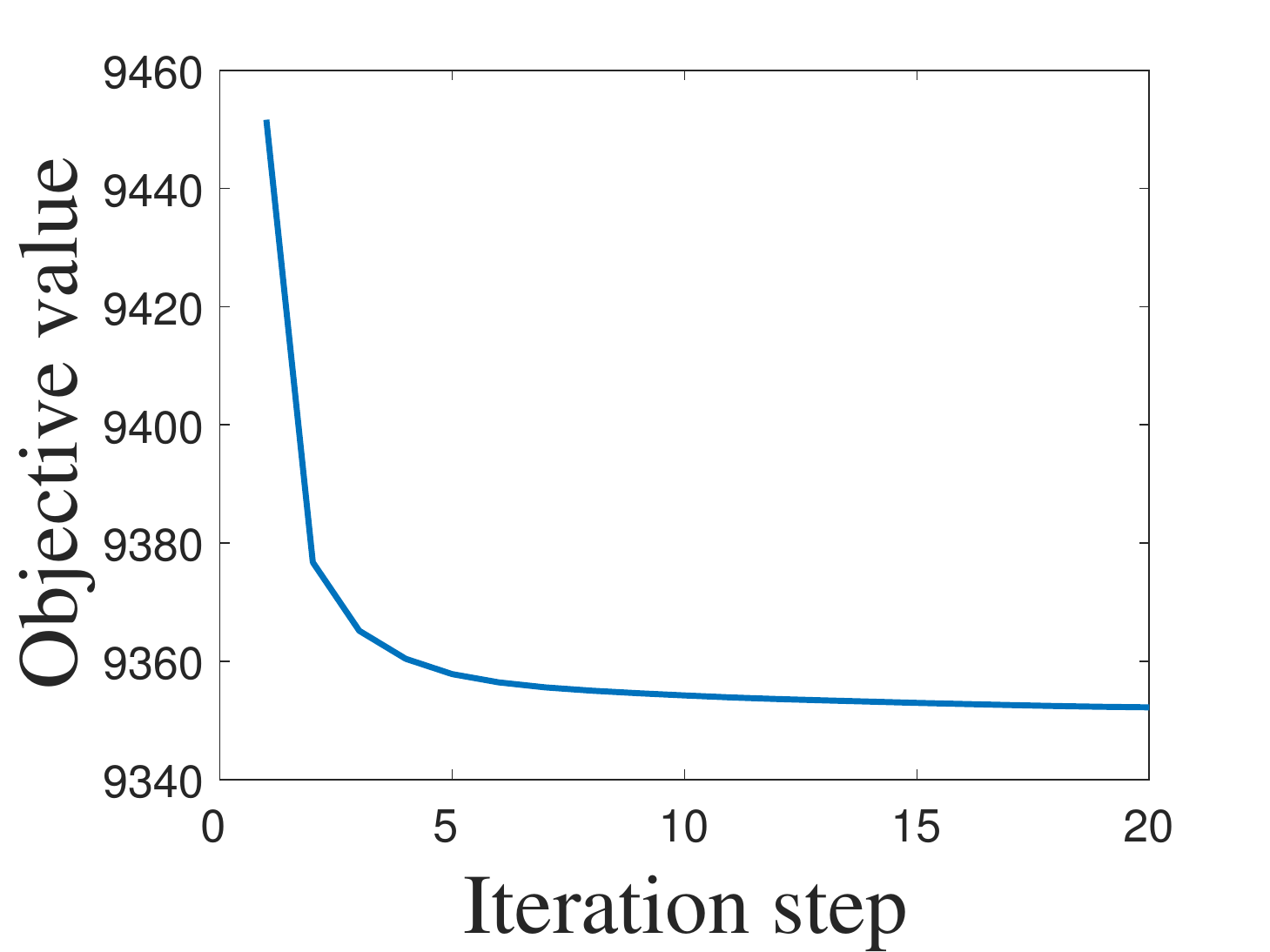}
		}
		\subfigure[YTF-20]{
			\includegraphics[width=0.5\linewidth]{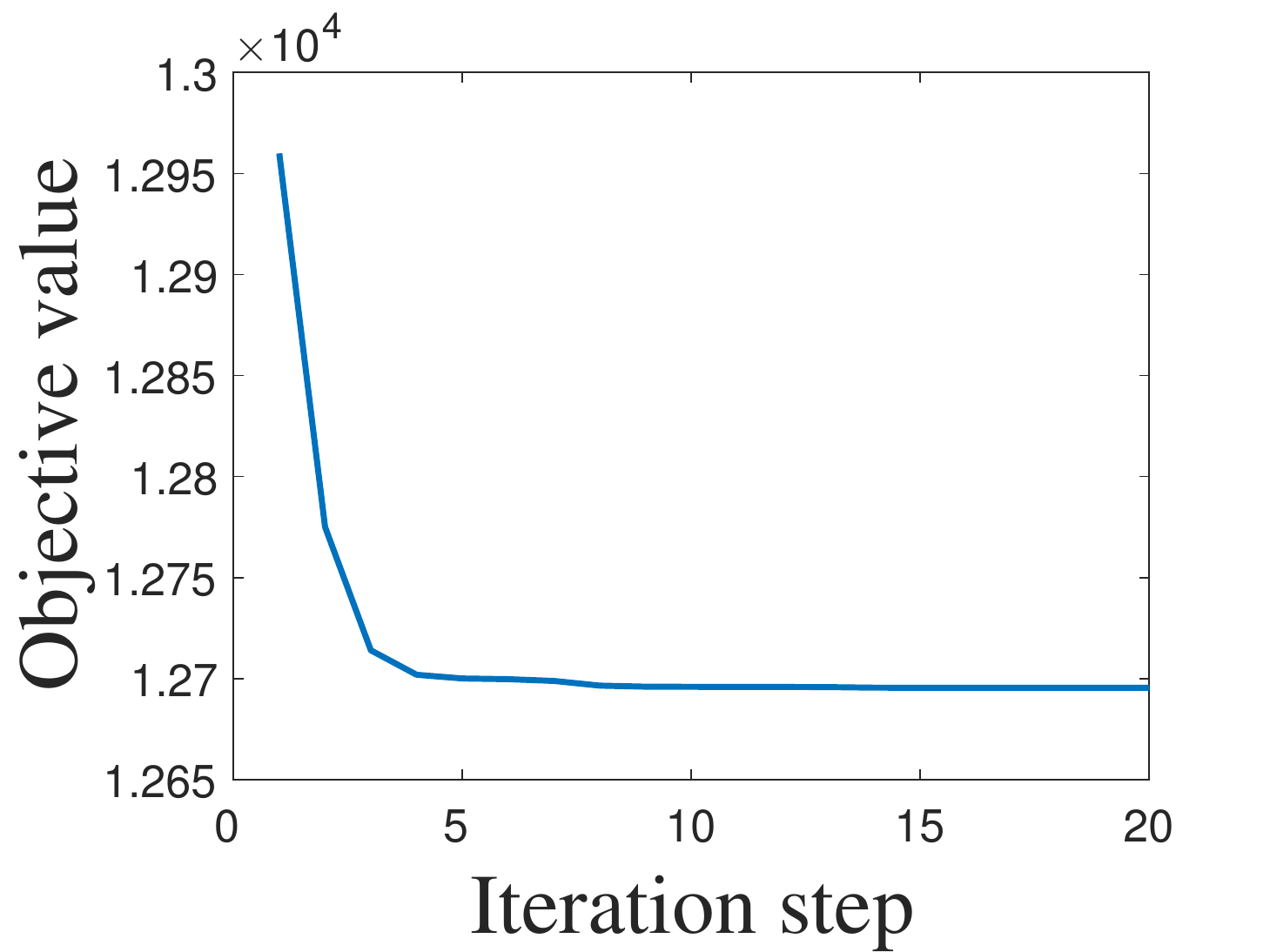}
		}
	}
	\caption{Convergence analysis: The objective value as a function of the iteration step on the four datasets.}
	\label{Figure:ConverAnalysis}
\end{figure}

\subsection{Parameter Analysis}

In this subsection, parameter analysis about the dimension of the view generation model $d$ is conducted on multiple datasets. Due to the page limit, the influences of different parameter values in terms of ACC and NMI on the four randomly selected benchmark datasets, i.e., HandWritten, SUNRGBD, VGGFace2-50 and YTF-20, are illustrated in \figurename~\ref{Figure:paraAnalysis}, in which the selection of $d$ is searched from $\left[k:5:300\right]$ according to different properties of multiple datasets. From the figure, it can be observed that the performance of AIMC is relatively stable over the corresponding ranges of parameter value on different datasets. Meanwhile, we can easily acquire the best clustering performance on different datasets according to the subfigures.

\subsection{Convergence Analysis}

Since the objective formulation in Eq.~\eqref{overallobj} solved by the alternate minimizing algorithm is non-increasing with iterations, the method can be guaranteed to converge finally. In this subsection, empirical convergence analysis is conducted on the HandWritten, SUNRGBD, VGGFace2-50 and YTF-20 datasets to verify the convergence property of the proposed AIMC method. The objective values on the four benchmark datasets are illustrated in \figurename~\ref{Figure:ConverAnalysis}. From the figure, we can observe that the corresponding objective value decreases monotonically to a minimum within the first 10 iterations, showing that the proposed model is able to converge within just a few iterations.

\section{Conclusion}
\label{sec:conclusion}
In this paper, we develop a novel model termed Adaptively-weighted Integral Space for Fast Multiview Clustering (AIMC) with nearly linear complexity. Within the framework, a view-specific generation model is designed to map the latent integral space into the corresponding view observation with diverse adaptive contributions. Reversely, with input view observations, the latent integral space can be restored by mapping back from given view observations with corresponding self-conducted confidences. In addition, the orthogonal centroid representation and cluster assignment are seamlessly constructed to approximate the latent integral space. By merging the reconstruction error and distortion error of data partition, AIMC is able to simultaneously discover the latent integral space as well as cluster partition. An alternate minimizing algorithm is developed to solve the optimization problem, which is proved to have linear time complexity \textit{w.r.t.} the sample size. Compared with the state-of-the-art multiview clustering methods and large-scale oriented methods, extensive experimental results on several large-scale datasets have validated the superiority of the proposed method.

\ifCLASSOPTIONcaptionsoff
  \newpage
\fi

\appendices

\bibliographystyle{IEEEtran}

\bibliography{sample-base}

\end{document}